\documentclass[conference]{IEEEtran}
\usepackage{graphicx}
\graphicspath{ {./figures/} }
\usepackage{amsmath,amssymb}
\usepackage[ruled,vlined]{algorithm2e}
\usepackage[english]{babel}
\usepackage{lipsum}  
\usepackage{hyperref}
\usepackage{cite}  
\usepackage{siunitx}
\usepackage{booktabs}
\begin{document}

\title{Assessment of Reward Functions in Reinforcement Learning for Multi-Modal Urban Traffic Control under Real-World limitations}

\author{\authorblockN{Alvaro Cabrejas Egea}
\authorblockA{MathSys Centre for Doctoral Training,\\
University of Warwick \& Vivacity Labs, London, UK\\
Email: a.cabrejas-egea@warwick.ac.uk}
\and
\authorblockN{Colm Connaughton}
\authorblockA{Warwick Mathematics Institute\\
University of Warwick\\
Email: c.p.connaughton@warwick.ac.uk}
}

\maketitle

\begin{abstract}
Reinforcement Learning is proving a successful tool that can manage urban intersections with a fraction of the effort required to curate traditional traffic controllers.
However, literature on the introduction and control of pedestrians to such intersections is scarce.
Furthermore, it is unclear what traffic state variables should be used as reward to obtain the best agent performance.
This paper robustly evaluates 30 different Reinforcement Learning reward functions for controlling intersections serving pedestrians and vehicles covering the main traffic state variables available via modern vision-based sensors.
Some rewards proposed in previous literature solely for vehicular traffic are extended to pedestrians while new ones are introduced.
We use a calibrated model in terms of demand, sensors, green times and other operational constraints of a real intersection in Greater Manchester, UK.
The assessed rewards can be classified in 5 groups depending on the magnitudes used: queues, waiting time, delay, average speed and throughput in the junction.
The performance of different agents, in terms of waiting time, is compared across different demand levels, from normal operation to saturation of traditional adaptive controllers.
We find that those rewards maximising the speed of the network obtain the lowest waiting time for vehicles and pedestrians simultaneously, closely followed by queue minimisation, demonstrating better performance than other previously proposed methods.
\end{abstract}

\IEEEpeerreviewmaketitle

\section{Introduction}
Effective traffic signal control is one of the key issues in Urban Traffic Control (UTC), effectively deciding how the available resources (green time) in our urban travel networks are allocated.
The efficiency associated with this allocation has an important impact on travel times, harmful emissions and economic activity.

First, fixed time controllers, and later, adaptive systems have been used to further optimise the global traffic flow in our cities.
Recent improvements in CPU and especially GPU power are allowing for vision-based sensors to gather large amounts of real-time data that a few years ago seemed unattainable, such as individual vehicle position and speeds, at a much lower marginal cost than would be feasible with traditional actuated sensors.
As a side effect of these developments the area covered by sensors is ever increasing, also becoming possible to direct some of these towards pedestrians.
This has allowed the development of novel smart control approaches, using real-time data to deliver cheap and responsive systems that can adapt to a variety of situations.
Reinforcement Learning (RL) approaches have been showing promising results in this field.
However, most of the existing works restrict themselves to vehicles only, not attempting to jointly optimise vehicular and pedestrian travel times, even though pedestrians are present in the great majority of real urban intersections.

This paper compares the performance of 30 different reward functions used by Deep Q-Network agents, split into 5 different classes based on the magnitudes they use, when controlling a simulation of a real-world junction in Greater Manchester (UK) that has been calibrated using 3.5 months of data gathered from Vivacity Labs vision-based sensors.

The paper is structured as follows:
Section \ref{lit} reviews previous literature in the field.
Section \ref{problem} states the mathematical framework used and provides some theoretical background.  
Section \ref{methods} reviews the environment, the agents and their implementation.
Section \ref{rewards} introduces the reward functions tested in this paper and provides their analytical expressions.
Section \ref{experiments} contains details about the training and evaluation of the agents.
Lastly, Section \ref{results} provides the experimental results and discusses them.
\section{Related Work}
\label{lit}
RL for UTC has been previously explored and discussed in a variety of research, aiming to eventually substitute existing adaptive control methods such as SCOOT\cite{scoot}, MOVA\cite{mova} and SCATS\cite{scats}.
The field has evolved from early inquiries about its theoretical potential use \cite{wiering2000} \cite{abdul2003} \cite{pra2010} \cite{abdulhai2010} \cite{abdoos2011}, to progressively more applied and realistic scenarios that look towards real-world use and deployment.
Recent works use different magnitudes in the reward function of the controlling agents (delay, queues, waiting time, throughput, ...), however, it is not clear what benefits are provided from choosing which.
The different magnitudes used as reward are thoroughly indexed in \cite{yau} \cite{survey2020} \cite{survey2020wei}, although no direct performance comparisons are made.
Different methods are taken regarding inputs, such as pixel-based vectors passed to a CNN \cite{liang2017} \cite{gao2017} \cite{mousavi2017}, per-lane state signals using fully connected neural networks \cite{survey2014} \cite{aslani2019} \cite{genders2019}, or hybrid approaches \cite{genders2016} \cite{gendersthesis} \cite{wan2018}.
Recent research suggests that more complex state representations only provide marginal gains, if any\cite{gendersstate}, so in this paper the second approach is taken.
A common thread in most previous works is the need for approximations about the network being studied and the lack of pedestrian modelling and joint optimisation for vehicles and pedestrians travel times.
As indicated in \cite{survey2020}, pedestrian implementation has a high impact on learning performance, being often discarded as unimportant or left for future work save for two exceptions \cite{geneticped} \cite{liu2017}, the first of which uses a genetic algorithm instead of RL, and the second explores a single reward function.
In this paper we attempt to cover this gap in the literature, providing a robust performance assessment of RL agents serving both vehicles and pedestrians, using a variety of rewards, both novel and from the literature, attempting to uncover what state variables should be used in the reward to obtain the best performance.
These are applied to a RL agent in a calibrated model of a real-world junction, using real geometry, calibrated demand, realistic sensor inputs and emulated traffic light controllers, to which some of these agents have been deployed to control real traffic in it since these experiments took place.
This paper delivers the future work deferred from \cite{previous} in terms of shifting the focus towards pedestrians and multi-objective optimisation, while keeping the problem grounded in the real world.
\section{Problem Definition}
\label{problem}
\subsection{Markov Decision Processes and Reinforcement Learning}
The problem is framed as a Markov Decision Process (MDP), satisfying the Markov property: given a current state $s_t$, the next state $s_{t+1}$ is independent of the succession of previous states $\{s_{t-1}, s_{t-2}, ..., s_0\}$.
An MDP is defined by the 5-element tuple:
\begin{enumerate}
\item The set of possible states $\mathcal{S}, s_i\in \mathcal{S}$.
\item The set of possible actions $\mathcal{A}, a_i\in \mathcal{A}$.
\item The probabilistic transition function between states $\mathcal{T}$.
\item The discount factor $\gamma \in [0,1]$ 
\item The scalar Reward Function $\mathcal{R}$. 
\end{enumerate}

The objective of an MDP optimisation is to find an optimal policy $\pi^*$, mapping states to actions, that maximises the sum of the expected discounted reward,
\begin{equation}
R_t = \mathbb{E} \bigg{[} \sum^{\infty}_{i=0} \gamma^i r_{t+i} \bigg{]} .
\label{eq:reward}
\end{equation}
In the case of RL for UTC, $\mathcal{T}$ is unknown, making it necessary to approach it from a model-free RL perspective.
Model-Free RL is an sub-field of RL covering how independent agents can take sequential decisions in an unknown environment and learn from their interactions in order to obtain $\pi^*$. 
There are two main approaches: Policy-Based RL, which maps states to a distribution of potential actions, and Value-Based RL, which is used in this paper and estimates the {\em value} (expected return) of the state-action pairs under a given policy $\pi$ defined  as
\begin{equation}
V^{\pi}(s) = \mathbb{E} [R_t|s,\pi].
\label{eq:value}
\end{equation}

\subsection{Q Learning and Value-Based RL}
Q-Learning\cite{watkins} is an off-policy model-free value-based RL algorithm. For any finite MDP, it can find an optimal policy which maximises expected total discounted reward, starting from any state\cite{melo}.
Q-Learning aims to learn an optimal action-value function $Q^*(s,a)$, defined as the total return after being in state $s$, taking action $a$ and then following policy $\pi^*$. 
\begin{equation}
Q^*(s,a) = \max_{\pi} \mathbb{E} \big[ R_t | s=s_t, a=a_t, \pi^* \big]
\label{eq:qlearning}
\end{equation}
Traditional table-based Q-Learning approximates $Q^*(s,a)$ recursively through successive Bellman updates,
\begin{equation}
Q^{\pi}(s_t,a_t) \leftarrow Q^{\pi}(s_t,a_t) + \alpha \big( y_t - Q(s_{t+1},a) \big)
\label{eq:bellmanupdate}
\end{equation}
with $\alpha$ the learning rate and $y_t$ the Temporal Difference (TD) target for the Q-function:
\begin{equation}
y_t = R_t + \gamma \,\,\, \max_{a_{t+1}} Q^{\pi}(s_{t+1},a_{t+1})
\end{equation}

This table representation is not useful for high dimensional cases, since the size of our table would increase exponentially, nor for continuous cases, since every distinct $s\in\mathcal{S}$ would require an entry.


\subsection{Deep Q Network}
One way of addressing the issues of Q-Learning in high dimensional spaces is to use neural networks as function approximators. This approach is called Deep Q-Network (DQN) \cite{mnih2015}.
The Q-function approximation is denoted then in terms of the parameters $\theta$ of the DQN as $Q(s,a,\theta)$.
DQN stabilises the learning process by introducing a Target Network that works alongside the main network. 
The main network with parameters $\theta$, approximates the Q-function, and the target network with parameters $\theta^-$  provides the TD targets for the DQN updates. 
The target network is updated every number of episodes by copying the weights $\theta^- \leftarrow \theta$. With $Q^{\pi} (s_{t+1}, a_{t+1}, \theta^-)$ representing the target network, it results in a TD target to approximate:
\begin{equation}
y_t = R_t + \gamma \,\,\, \max_{a_{t+1}} Q^{\pi} (s_{t+1}, a_{t+1}, \theta^-).
\end{equation}
\section{Methods}
\label{methods}
\subsection{Reinforcement Learning Agent}
The agent used to obtain these results is a standard implementation of a DQN in PyTorch \cite{pytorch}, optimising its weights via Stochastic Gradient Descent \cite{kiefer} using ADAM \cite{adam} as optimizer.
The learning rate is $\alpha=10^{-5}$ and the discount factor is $\gamma = 0.8$ for all simulations.
The Neural Network in the agent uses 2 hidden, fully connected layers of sizes 500 and 1000 respectively, using ReLU as an activation function.
\begin{algorithm}
\SetAlgoLined
 Create main network with random weights $\theta$\;
 Create target network with random weights $\theta^-$\;
 Create replay memory $M$ with capacity $L$\;
 Define frequency $F$ for copying weights to target network\;
 \For{each episode}{ 
    measure initial state $s_0$\;
    \While{episode not done}{
        select action $a_t$ according to $\epsilon$-greedy policy\;
        implement $a_t$\;
        advance simulation until next action is needed\;
        measure new state $s_{t+1}$, and calculate reward $R_{t+1}$\;
        store transition tuple $(s_t,a_t,R_{t+1},s_{t+1})$ in $M$\;
        $s \leftarrow s_{t+1}$\;
        }
    $b \leftarrow$ sample minibatch of transitions tuples from $M$\;
    \For{each transition $x_i = (s_i,a_i,R_{i+1},s_{i+1})$ in $b$}{
        $y_i = R_{i+1} + \gamma \max_{a} Q(s_{i+1}, a', \theta^-)$
        }
    Stochastic Gradient Descent on $\theta$ over all $(x_i,y_i) \in b$\;
    \If{number of episode is multiple of $F$}{
        $\theta^- \leftarrow \theta$
        }
    }
 \caption{Schematic Learning Process}

\end{algorithm}
\subsection{Reinforcement Learning Environment}
The environment is modelled in the microscopic traffic simulator SUMO \cite{sumo}, representing a real-world intersection in Greater Manchester, UK.
The junction consists of four arms, with 6 incoming lanes (two each in the north-south orientation, and one each in the east-west orientation) and 4 pedestrian crossings.
The real-world site also contains 4 Vivacity vision-based sensors, able to supply occupancy, queue length, waiting time, speed and flow data.
The demand and turning ratios at the junction have been calibrated using 3.5 months of journey time and flow data collected by these sensors.
The environment includes an emulated traffic signal controller, responsible for changing between the different stages in the intersection and enforcing the operational limitations, which are focused on safety.
This includes enforcing green times, intergreen times, as well as determining allowed stages.  
A stage is defined as a group of non-conflicting green lights (phases) in a junction which change at the same time.
The agent decides which stage to select next and requests this from an emulated traffic signal controller, which moves to that stage subject to its limitations, which are primarily safety-related.
The data available to the agent is restricted to what can be obtained from the sensors.     
\begin{figure}                                                
\centering                                                    
\includegraphics[width=2.5in]{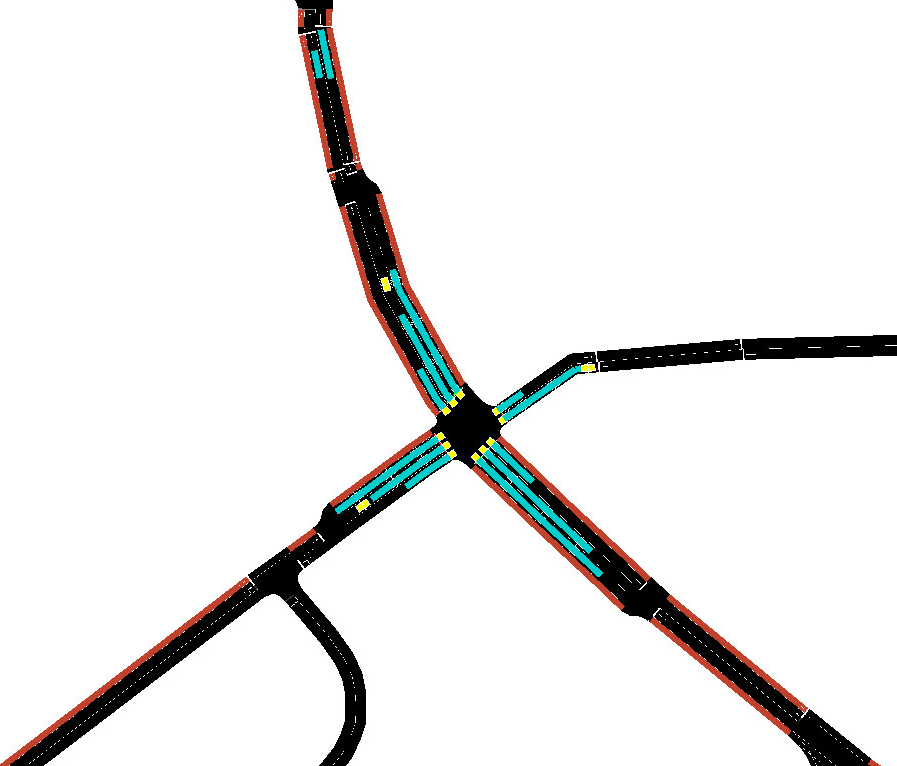}                                    
\caption{Study Junction model in SUMO with a schematic representation of the areas covered by vision-based sensors.}                                  
\label{intersection}                                               
\end{figure}     
\subsection{State Representation}
The agent receives an observation of the simulator state as input, using the same state information across all experiments here presented.
Each observation is a combination of the state of the traffic controller (which stage is active) and data from the sensors.
The data from the sensors is comprised of the occupancy in each lane area detector and a binary signal representing whether the pedestrian crossing button has been pushed.
The agent receives a concatenation of the last 20 measurements at a time, covering the previous 12 seconds at a resolution of 0.6 seconds.

\subsection{Actions of the Agent}
The junction is configured to have 4 available stages. 
The agent is able to choose Stage 2, Stage 3 or Stage 4, yielding an action space size of 3.
Stage 1 services a protected right turn coming from the north. It is used by the traffic light controller, as a transitional step for reaching Stage 2, as defined by the transport authority.
Stage 2 deals with the traffic in the north-south orientation.
Stage 3 is the pedestrian stage, setting all pedestrian crossings to green, and all other phases to red.
Stage 4 services the roads in the east-west orientation, which have considerable demand.

Once the controller has had a stage active for the minimum green time duration, the agent is requested to compute the value of all potential state-action pairs (i.e. the value of other stages given the current state) once per time-step.
From these, the action with the highest expected value is selected following an $\epsilon$-greedy policy\cite{suttonbarto}.
Should the agent choose the same action, the current stage will be extended for a further time-step (0.6 seconds).
There is no built-in limit to the maximum number of said extensions, leaving it for the agent to learn the optimal green time for any given situation.
If a different stage is chosen, then the controller will proceed to the intergreen transition between them.

There are 2 situations that further add to the complexity of this control process:
\begin{enumerate}
\item Variable number of extensions, and hence length of the stages, creates a distribution of values over the state-action pairs in most rewards, which the agent must approximate. The variance of this distribution will be higher than the variance that would be obtained using constant stage length.
\item The requirement that Stage 1 must be used as an intermediate step to reach Stage 2 implies less certainty in the control process than in other stages, since there is an unaccounted dilated temporal horizon between the state that triggered the action, and the effects of said action over the state variables.
\end{enumerate}

\begin{figure}                                                
\centering                                                    
\includegraphics[width=2.5in]{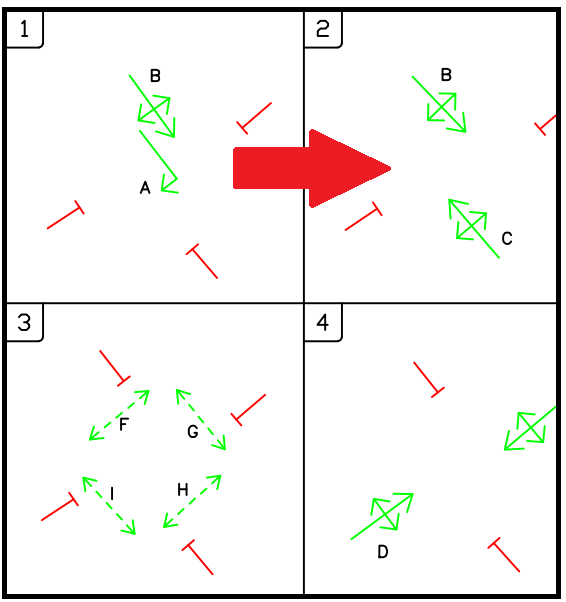}                                    
\caption{Allowed stages and the phases that compose them. Stage 1 is an intermediate Stage, which is necessary to go through to reach Stage 2.}                                  
\label{intersection_phases}  
                                            
\end{figure}     
\subsection{Modal Prioritisation and Adjusting by Demand}
The agent serves vehicles and pedestrians arriving at the intersection, seeking to jointly optimise the intersection for both modes of transport.

All the reward functions presented in this paper follow the same structure.
The reward, as seen by the agent, will be a linear combination of an independently calculated reward for the vehicles and another for the agents, as it can be seen in Eq. \ref{eq:modal}.
\begin{equation}
R_t = a * R^v_t + b * R^p_t;\,\,\,\,\,\,\ a+b = 1
\label{eq:modal}
\end{equation}
In this way, $a$ and $b$ are the Modal Prioritisation coefficients for our rewards, with $R^v, R^p$ being respectively the vehicular and pedestrian rewards.

Of the rewards presented in the following section, those that were more sensitive towards the relative ratio of the demand between pedestrian and vehicles require manual tuning of the modal prioritisation parameters.
While undesirable from a modeller and operator point of view since it partially counters the benefits that RL provides in terms of self-adjustment, they are provided so potential users and researchers can evaluate the trade-offs between potential increased performance and increased configuration effort.
The mentioned series will be identified by the weight applied to the pedestrians.
As such, series identified as P80 and P95 represent those in which the weights were $a=0.2$, $b=0.8$, and $a=0.05$, $b=0.95$ respectively. 
Those series without an identifier did not require modal prioritisation ($a = b$).

Another addition that can be made to the rewards is to add a term scaling the difficulty with the demand level, implicitly accepting that higher demand typically worsens the performance of a network, independent of the actions of the controlling agent.
These series are identified with the suffix AD (Adjusted by Demand).
\section{Reward Functions}
\label{rewards}
All reward functions tested are presented in this section with their analytical expressions.

Let $N$ be the set of lane queue sensors present in the intersection.
Let $M$ be the set of pedestrian occupancy sensors in the junction.
Let $V_t$ and $P_t$ be respectively the set of vehicles in incoming lanes, and the set of pedestrians waiting to cross in the intersection at time $t$. 
Let $s_v$ be the individual speeds of the vehicles, $\tau^v$ and $\tau^p$ the waiting times of vehicles and pedestrians, respectively.
Let $\rho_v$ and $\rho_p$ be the vehicular and pedestrian flows across the junction over the length of the action.
Let $t^p$ be the time at which the previous action was taken and $t^{pp}$ the time of the action before that.
Lastly, let $t^v_e$ and $t^p_e$ be the entry times of vehicles and pedestrians to the area covered by sensors. 

\subsection{Queue Length based Rewards}
\subsubsection{Queue Length}
Similar to \cite{pra2010}, used in \cite{aslani2019}, the reward is the negative sum at $t$ of queues ($q$) over all ($n,m$) sensors.
\begin{equation}
    R_t = - \sum_{n \in N} q^v_{t} - \sum_{m \in M} q^p_{t}
\label{eq:queue}
\end{equation}

\subsubsection{Queue Squared}
As seen in \cite{gendersthesis}, this function squares the result of adding all queues.
\begin{equation}
   R_t = -  \bigg( \sum_{n \in N} q^v_{t} \bigg)^2 - \bigg( \sum_{m \in M} q^p_{t} \bigg)^2
\label{eq:queuesq} 
\end{equation}

\subsubsection{Queues PLN}
As Queue length, but dividing the sum by the phase length (Phase Length Normalisation), approximating the reward that the action generates by unit of time it is active.
\begin{equation}
    R_t = - \frac{1}{t-t^p}  \sum_{n \in N} q^v_{t} -  \sum_{m \in M} q^p_{t}
    \label{queuepln}
\end{equation}

\subsubsection{Delta Queue}
The reward is the variation of the sum of queues between actions.
\begin{equation}
    R_t =  \bigg( \sum_{n \in N} q^v_{t^p} -\sum_{n \in N} q^v_{t} \bigg) +  \bigg( \sum_{m \in M} q^p_{t^p} - \sum_{m \in M} q^p_{t} \bigg)
    \label{deltaqueue}
\end{equation}

\subsubsection{Delta Queue PLN}
As Delta Queue, but dividing the sum by the phase length (Phase Length Normalisation).
\begin{equation}
    R_t = - \frac{1}{t-t^p} \bigg(  \big( \sum_{n \in N} q^v_{t^p} - \sum_{n \in N} q^v_{t} \big) - 
    \big(  \sum_{m \in M} q^p_{t^p}  - \sum_{m \in M} q^p_{t} \big) \bigg)
    \label{deltaqueuepln}
\end{equation}

\subsection{Waiting Time based Rewards}
These rewards require Modal Prioritisation weights.
\subsubsection{Wait Time}
The reward is the negative sum of time in queue accumulated since the last action by all vehicles. 
\begin{equation}
R_t = - \bigg(a \sum_{v \in V_t} \tau^v_{t} + b \sum_{p \in P_t} \tau^p_{t} \bigg)
\label{eq:wait_time}
\end{equation}

\subsubsection{Delta Wait Time}
As seen in \cite{liang2017}, the reward is the variation in queueing time between actions.
\begin{equation}
R_t = a \bigg( \sum_{v \in V_t} \tau^v_{t_p} -  \sum_{v \in V_t} \tau^v_{t} \bigg) + b \bigg( \sum_{p \in P_t} \tau^p_{t_p} -  \sum_{p \in P_t} \tau^p_{t} \bigg)
\label{eq:delta_wait_time}
\end{equation}

\subsubsection{Waiting Time Adjusted by Demand}
Negative sum of waiting time, adding a factor to scale it accordingly with an estimate of the demand ($\hat{d}$).
\begin{equation}
   R_t = -\frac{1}{\hat{d}} \bigg( a \sum_{v \in V_t} \tau^v_{t} + b \sum_{p \in P_t} \tau^p_{t} \bigg)
\label{eq:wait_time_norm} 
\end{equation}

\subsection{Delay based Rewards}
These rewards require Modal Prioritisation weights.
\subsubsection{Delay}
As seen in \cite{wan2018}. Negative weighted sum of the delay by all entities. Delay is understood as deviation from the maximum allowed speed. 
For the pedestrians, the time in queue is used given that, from the point of view of the sensors, pedestrian presence is binary.
Assuming a simulator time step of length $\delta$:
\begin{equation}
    R_t = - \bigg( a \sum_{v \in V_t}  \sum_{t^v_e}^t \delta \big( 1-\frac{s_v}{s_{max}} \big) + b \sum_{p \in P_t} \tau^p_t \bigg)
\label{eq:delay}
\end{equation}

\subsubsection{Delta Delay}
First seen in \cite{abdulhai2010} and used in \cite{genders2016} \cite{gao2017} \cite{mousavi2017} and \cite{gendersstate}.
The reward is the variation between actions of the delay as calculated in Eq. (\ref{eq:delta_wait_time}).
\begin{equation}
\begin{split}
R_t =  a \bigg( \sum_{v \in V_t} \sum_{t^{pp}}^{t^p} \delta & \big( 1-\frac{s_v}{s_{max}} \big) - \sum_{v \in V_t}  \sum_{t^p}^t \delta \big( 1-\frac{s_v}{s_{max}} \big) \bigg) \\
& + b \bigg( \sum_{p \in P_t} \tau^p_{t^p} - \sum_{p \in P_t} \tau^p_{t} \bigg) 
\end{split}
\label{eq:changedelay} 
\end{equation}

\subsubsection{Delay Adjusted by Demand}
Same as in Eq. (\ref{eq:delay}), introducing a scaling demand term.
\begin{equation}
    R_t = -\frac{1}{\hat{d}} \bigg( a \sum_{v\in V_t}  \sum_{t^p}^t \delta \big( 1-\frac{s_v}{s_{max}} \big) + b \sum_{p \in P_t} \tau^p_{t} \bigg)
\label{eq:delay_ad}
\end{equation}

\subsection{Average Speed based Rewards}
\subsubsection{Average Speed, Wait Time Variant}
The vehicle reward is the average speed of vehicles in the area covered by sensors and normalised by the maximum speed.
The pedestrian reward is the minimum between the sum of the waiting time of the pedestrian divided by a theoretical desirable maximum waiting time $\tau_{max}$ and 1.
This produces two components of the reward $R_p, R_v \in [0,1]$.

\begin{equation}
    R_t =  \frac{\sum_{v \in V_t} \frac{s}{s_{max}}}{\sum_{V_t} v}  +  \min \big( \sum_{p \in P_t} \frac{\tau^p_{t}}{\tau_{max}} \, , 1 \big)
\label{eq:avgspeed_wait}
\end{equation}

\subsubsection{Average Speed, Occupancy Variant}
Vehicle reward as in the previous entry.
Pedestrian reward is the minimum between the sum of pedestrians waiting divided by a theoretical maximum desirable capacity $p_{max}$ and 1. 
\begin{equation}
    R_t =    \frac{ \sum_{v \in V_t} \frac{s}{s_{max}}}{\sum_{V_t} v} +  \min \big( \sum_{p \in P_t} \frac{p}{p_{max}} \, , 1 \big)
\label{eq:avgspeed_occ}
\end{equation}

\subsubsection{Average Speed Adjusted by Demand, Demand and Occupancy Variants}
As in the previous two entries, adding a multiplicative factor equal to the estimation of the demand $\hat{d}$, scaling the reward with the difficulty of the task.

\subsection{Throughput based Rewards}
These rewards require Modal Prioritisation weights.
\subsubsection{Throughput}
The reward is the sum of the pedestrians and vehicles that cleared the intersection since the last action.
\begin{equation}
    r_t = a \sum_{t_p}^t \rho_v + b \sum_{t_p}^t \rho_p
\label{eq:throughput}
\end{equation}
\begin{figure*}                                                
\centering                                                    
\includegraphics[width=\textwidth]{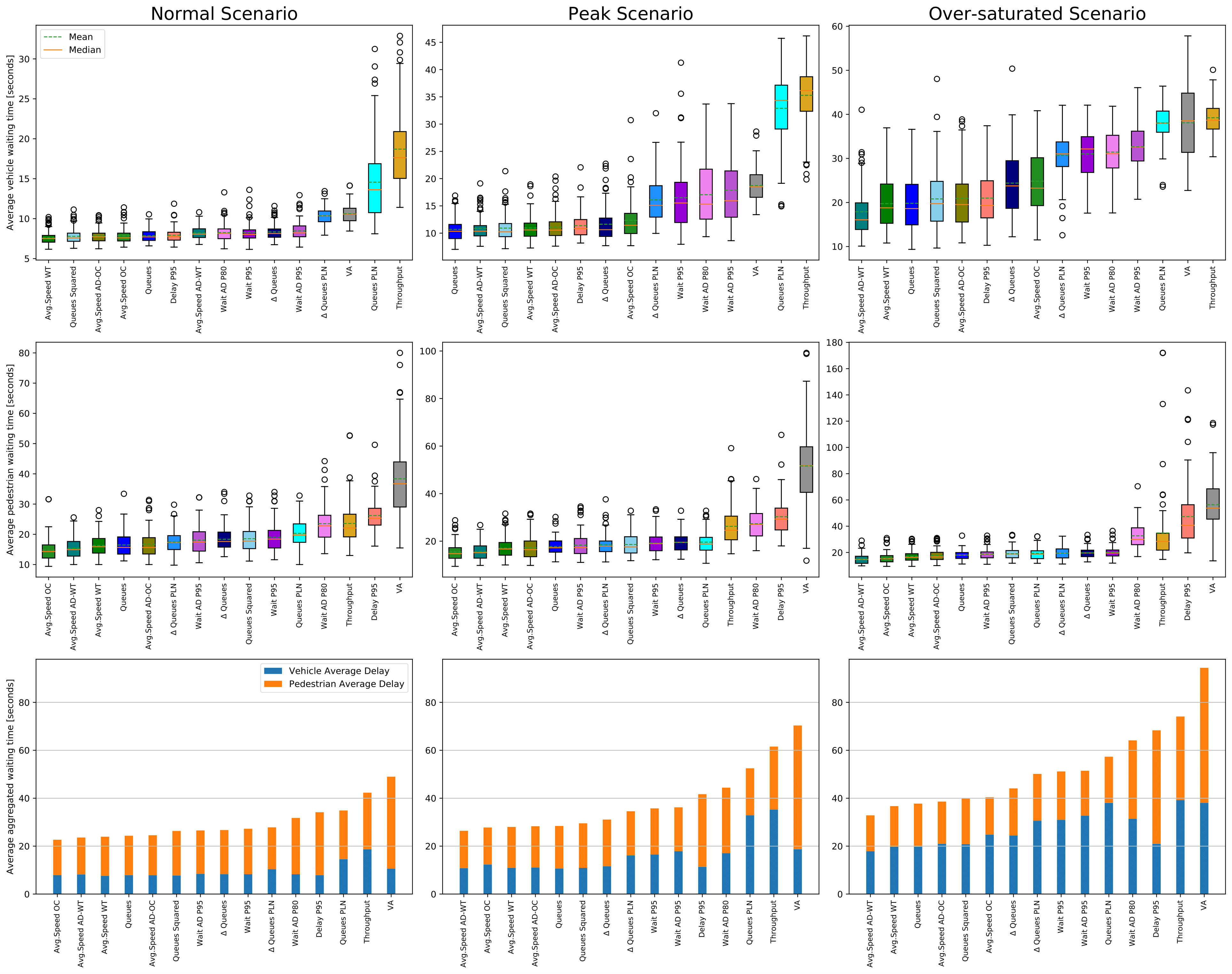}                                    
\caption{Top row: Average vehicular waiting time distribution for the fifteen best performing agents across demand levels. Middle row: Average pedestrian waiting time distribution across demand levels. Bottom row: Aggregated vehicular and pedestrian mean performance across demand levels.}                                  
\label{fig:results_9_grid}    
                                          
\end{figure*}

\begin{table*}[!htp]
\centering
\setlength\tabcolsep{0pt}

\caption{Average waiting time in seconds for all agents across demand levels}
\label{tab:some example table}

\begin{tabular*}{\textwidth}{
  @{\extracolsep{\fill}}
  l
  S[table-format=-1.2(1)]
  S[table-format=-1.2(1)] 
  S[table-format=-1.2(1)]
  S[table-format=-1.2(1)]
  S[table-format=-1.2(1)]
  S[table-format=-1.2(1)]
  @{}
}
\toprule
& \multicolumn{2}{c}{Normal Scenario} &
\multicolumn{2}{c}{Peak Scenario} &
\multicolumn{2}{c}{Oversaturated Scenario} \\

Scenario & {Vehicles} & {Pedestrians} & {Vehicles} & {Pedestrians}  & {Vehicles}  & {Pedestrians}\\
\midrule
Queues                 & 7.87\pm 0.83 & 16.47\pm 3.95 & 10.68\pm 2.06 & 17.73\pm 3.64 & 19.80\pm 6.01 & 17.94\pm 3.48  \\
Queues Sq.             & 7.79\pm 0.93 & 18.55\pm 4.47 & 10.92\pm 2.41 & 18.60\pm 4.81 & 20.80\pm 6.88 & 19.02\pm 4.38  \\
Queues PLN             & 14.57\pm 4.91 & 20.31\pm 4.94 & 32.90\pm 6.36 & 19.59\pm 4.78 & 38.04\pm 3.93 & 19.28\pm 4.87  \\
$\Delta$ Queues        &  8.34\pm 1.04 & 18.37\pm 3.94 & 11.63\pm 3.09 & 19.45\pm 3.75 & 24.40\pm 7.20 & 19.70\pm 3.80  \\
$\Delta$ Queues PLN    & 10.37\pm 1.10 & 17.45\pm 3.59 & 16.11\pm 4.38 & 18.44\pm 4.45 & 30.64\pm 5.00 & 19.49\pm 4.32  \\
\midrule
Average Speed - Wait     & 7.61\pm 0.84 & 16.31\pm 3.82 & 10.94\pm 2.19 & 17.05\pm 4.67 & 10.62\pm 1.17 & 38.36\pm 12.66  \\
Average Speed - Occ      & 7.86\pm 0.94 & 14.79\pm 3.97 & 12.34\pm 3.44 & 15.43\pm 3.84 & 24.84\pm 7.31 & 15.56\pm 4.49  \\
Average Speed AD - Wait  & 8.20\pm 0.80 & 15.37\pm 3.48 & 10.85\pm 2.11 & 15.55\pm 3.84 & 17.89\pm 5.68 & 14.95\pm 3.80  \\
Average Speed AD - Occ   & 7.85\pm 0.88 & 16.68\pm 4.83 & 11.10\pm 2.44 & 17.20\pm 4.93 & 20.93\pm 6.83 & 17.66\pm 5.01  \\
\midrule
Wait Time              & 7.80\pm 0.90 & 41.05\pm 19.40 & 14.65\pm 4.73 & 110.34\pm 59.56 & 28.82\pm 4.83 & 228.46\pm 159.81  \\
Wait Time P80          & 8.20\pm 1.26 & 28.80\pm 9.05 & 14.94\pm 4.81 & 54.29\pm 35.00 & 30.01\pm 4.84 & 113.68\pm 52.00  \\
Wait Time P95          & 8.26\pm 1.15 & 19.00\pm 4.67 & 16.51\pm 5.94 & 19.24\pm 4.44 & 31.02\pm 5.26 & 20.14\pm 4.16  \\
\midrule
Wait Time AD           & 7.83\pm 0.99 & 56.00\pm 30.22 & 14.84\pm 4.84 & 169.11\pm 92.44 & 27.52\pm 5.01 & 324.12\pm 212.37  \\
Wait Time AD P80       & 8.25\pm 1.13 & 23.52\pm 5.73 & 17.05\pm 5.91 & 27.35\pm 6.29 & 31.43\pm 4.95 & 32.69\pm 9.67  \\
Wait Time AD P95       & 8.48\pm 1.19 & 18.07\pm 4.75 & 17.88\pm 6.25 & 18.30\pm 5.02 & 32.67\pm 5.28 & 18.78\pm 4.37  \\
\midrule
$\Delta$ Wait Time     & 9.12\pm 1.23 & 82.57\pm 36.55 & 15.28\pm 5.09 & 326.07\pm 175.84 & 24.16\pm 6.77 & 594.03\pm 273.64  \\
$\Delta$ Wait Time P80 & 8.94\pm 1.38 & 33.35\pm 17.34 & 16.68\pm 4.65 & 81.64\pm 49.48 & 30.38\pm 4.50 & 149.79\pm 105.07  \\
$\Delta$ Wait Time P95 & 10.02\pm 1.66 & 42.36\pm 16.59 & 16.27\pm 5.33 & 72.27\pm 44.89 & 26.88\pm 6.22 & 174.85\pm 109.01  \\
\midrule
Delay                  & 6.39\pm 0.40 & 849.52\pm 318.33 & 8.59\pm 0.89 & 849.52\pm 318.33 & 14.43\pm 3.16 & 849.52\pm 318.33  \\
Delay P80              & 8.39\pm 1.10 & 46.78\pm 16.52 & 11.52\pm 2.36 & 78.91\pm 35.73 & 20.93\pm 7.02 & 143.27\pm 72.39  \\
Delay P95              & 7.92\pm 0.89 & 26.21\pm 4.86 &  11.38\pm 2.42 & 30.30\pm 7.64 & 20.99\pm 6.29 & 47.34\pm 23.27  \\
\midrule
Delay AD               & 6.71\pm 0.43 & 811.38\pm 352.38 & 8.79\pm 0.96 & 811.38\pm 352.38 & 14.08\pm 3.31 & 811.38\pm 352.38  \\
Delay AD P80           & 7.74\pm 0.81 & 44.55\pm 17.51 & 10.68\pm 1.95 & 122.05\pm 112.18 & 18.92\pm 6.46 & 404.54\pm 252.47  \\
Delay AD P95           & 7.83\pm 0.84 & 48.76\pm 24.28 & 11.62\pm 3.02 & 180.59\pm 123.18 & 21.77\pm 6.82 & 425.35\pm 234.33  \\
\midrule
$\Delta$ Delay         & 11.18\pm 2.93 & 211.41\pm 116.86 & 26.98\pm 6.75 & 546.51\pm 263.71 & 34.97\pm 3.45 & 393.81\pm 267.95  \\
$\Delta$ Delay P80     & 10.62\pm 2.34 & 66.46\pm 30.32 & 20.51\pm 6.04 & 180.64\pm 107.80 & 29.70\pm 4.97 & 307.76\pm 218.11  \\
$\Delta$ Delay P95     & 8.23\pm 1.29 & 99.40\pm 59.76 & 15.22\pm 4.97 & 221.92\pm 133.24 & 25.35\pm 6.43 & 398.13\pm 240.03  \\
\midrule
Throughput             & 18.71\pm 4.79 & 23.60\pm 6.88 & 35.28\pm 5.60 & 26.26\pm 8.14 & 39.24\pm 3.72 & 34.86\pm 28.54  \\
Throughput P80         & 35.53\pm 10.87 & 51.96\pm 31.20 & 47.60\pm 5.99 & 65.91\pm 37.86 & 47.85\pm 5.15 & 84.93\pm 49.08  \\
Throughput P95         & 26.28\pm 8.81 & 101.07\pm 65.21 & 56.39\pm 10.72 & 130.98\pm 84.11 & 74.10\pm 13.94 & 74.46\pm 57.96  \\
\midrule
Vehicle Actuated System D  & 10.62\pm 1.17 & 38.36\pm 12.66 & 18.73\pm 2.92 & 51.62\pm 16.50 & 38.10\pm 8.26 & 56.32\pm 19.58  \\
Maximum Occupancy     & 6.92\pm 0.54 & 196.09\pm 130.04 & 10.02\pm 1.75 & 397.20\pm 213.06 & 21.57\pm 5.10 & 596.32\pm 253.80  \\
\bottomrule
\label{table}
\vspace{-7mm}
\end{tabular*}
    
\end{table*}
\section{Experiments}
\label{experiments}
\subsection{DQN Agents Training}
The training process covers 1500 episodes running for 3000 steps of length $\delta=0.6$ seconds for a simulated time of 30 minutes (1800 seconds).
The traffic demand is increased as the training advances, with the agent progressively facing sub-saturated, near-saturated and over-saturated scenarios, with a minimum of 1 vehicle / 3 seconds (1200 vehicles/h) and a maximum of 1 vehicle / 1.4 seconds (2571 vehicles/h).

For each reward function, 10 copies of the agent are trained, and their performance was compared against two reference systems. These are Maximum Occupancy (longest queue first) and Vehicle Actuated System D \cite{highways} (vehicle-triggered green time extensions), which is commonly used in the UK. The agent performing best against the reference systems in each class is selected for detailed scoring.
\subsection{Evaluation and Scoring}
Each selected agent is tested and its performance scored over 100 copies of 3 different scenarios with different demand levels. Each evaluation is the same length as the training episodes, with the demand kept constant during each run.
These three scenarios are aimed to test the agents during normal operation, peak times and over-saturated conditions, and will be henceforth referred to as Normal, Peak and Over-saturated Scenarios.
Peak Scenario uses the level of demand observed in the junction that results in saturated traffic conditions under traditional controllers.

The Normal Scenario uses an arrival rate of 1 vehicle / 2.1 seconds (1714 vehicles/h). Peak Scenario uses an arrival rate of 1 vehicle / 1.7 seconds (2117 vehicles/h). Over-saturated Scenario uses an arrival rate of 1 vehicle / 1.4 seconds (2400 vehicles/h)

\section{Results and Discussion}
\label{results}
The results from the simulations of the different reward functions are summarised in Fig. \ref{fig:results_9_grid}, including the performance of the 15 rewards found to have lower waiting times and seeming most desirable in practice. They are detailed for all 30 rewards in Table \ref{table}.
In Fig. \ref{fig:results_9_grid}, the distribution of pedestrian and vehicle waiting times, and the combination of mean performances for both modes of transportation across 100 repetitions of each demand level are presented.
Table \ref{table} shows the mean waiting time for each distribution and their standard deviation, also calculated across all three demand levels. 

The results display further evidence that RL agents can reach better performance than reference adaptive methods, more evidently so when pedestrians are added.
In the case of MO, the bad performance can be framed within the need of having more pedestrians queued than vehicles in any sensor in order to start the pedestrian stage.
VA suffers due to its predisposition towards extending green times by 1.5s in the presence of any vehicle, making it more difficult to reach a state in which the pedestrian stage can be started.
Both of these characteristics make the vanilla reference methods less suited for intersections including pedestrians than the RL methods presented in Fig. \ref{fig:results_9_grid}, especially in situations of high demand.

At a global level, methods based on maximisation of the average network speed show the lowest global waiting times for pedestrians and vehicles combined across all demand levels, while also obtaining some of the lowest spreads, as shown in the case with no pedestrians \cite{previous}.
Their performance is closely followed by Queue minimisation, which obtains the lowest average waiting times for vehicles in the Normal and Peak Scenarios, but falls behind in Over-saturated conditions and when dealing with pedestrians.
Queue Squared minimisation has a comparable yet slightly worse performance, followed by Delta Queues and Delta Queues PLN.
This last reward has shown to obtain better performance with higher demand, which is consistent with it generating less variance in the state, since it is modelling for arrival rates given an action, and makes it an option that could be further explored for permanently congested intersections.
Prioritised rewards based on Waiting Time show acceptable performance, but also a high sensitivity to the changes in the modal prioritisation weights. 
This is similar to the behaviour shown by the Delay-based rewards, which overall perform worse, potentially due to the need to use Wait Time for pedestrians, mixing the state variables, although this does not seem to be an issue for average speed based rewards.
Without a weight configuration heavily favouring the pedestrians, these reward functions were found to converge for vehicles only, obtaining the lowest vehicle waiting times overall in the case of the Delay functions, at the expense of rarely, if ever, serving pedestrians.
The suitability of a given choice of modal prioritisation weights is further affected by the functional form of the reward.
In the results, it can be observed that while in general the choice $(a=5,b=95)$ obtains better results (e.g. Wait Time and Delay), for certain functional choices the prioritisation $(a=20,b=80)$ is the one producing the best results, which would not be the case if the suitability of the weights was only affected by the relative demand ratios between vehicles and pedestrians. 
This is the case with Throughput based functions, which, unlike the Wait and Delay functions, obtained lower waiting times with equal modal weights, and a general wait time increase as the weights become more skewed towards the pedestrians.
Rewards using Differences in Delay or Wait Time, having good performance in the literature, were found either not to converge for pedestrians or to produce mediocre results.
The addition of a demand scaling term generates, in general, a slight improvement in waiting times across the rewards using Wait Time and Delay, particularly at higher demand levels. 

Overall, dominance shown by speed maximisation methods could be attributed to several factors.
Average Speed based functions, as Queue based functions, obtain an instantaneous snapshot of a magnitude that does not intrinsically grow over time, as opposed to Delay, Wait and Throughput, so it exclusively encodes information about the moment the action is requested.
It can also be argued that speed maximisation rewards are not affected by the correspondence between agent actions and time-steps in the environment. In the specific case of RL for UTC, the values of the reward received by the agent using a reward based on Queues, Delay, Wait or Throughput are a function of the length of the phase that generated them, making them theoretically less suitable for the underlying MDP than speed maximisation.
Lastly, speed maximisation and queue minimisation have an extra benefit that makes them into serious candidates for expansive real-world use: the lack of need for modal prioritisation tuning.
One of the main selling points of ML and RL methods stems from their ability to perform equal or better than traditional systems at a lower cost in a variety of situations.
However, a lengthy manual tuning process in order to find the exact weights for a given junction is not only untranslatable to any other intersection, but may also not result in reduced planning and execution times compared with traditional control.
The lack of need for manual tuning, especially in the case of Average Speed functions, which are specifically crafted to avoid this, make them in our view more applicable in a wider and faster manner than any of the other reward functions here presented.

One limitation of this paper is that the results are only relevant in the case of value-based DQN agents as introduced in Section \ref{problem} and Section \ref{methods}, and not for CNN or Policy Gradient architectures.
This work could be extended to account for other modes of transportation, performing a similar optimisation based on different vehicle classes (buses, cyclists, personal vehicles, trucks, etc.). The optimisation could seek to prioritise them based on different criteria (e.g. priority to cyclists and public transport during rush hours or weighting vehicles according to the expected number of passengers).

%

%

\section*{Acknowledgment}
This work was part funded by EPSRC Grant EP/L015374 and part funded by InnovateUK grant 104219. 
Vivacity Labs thanks Transport for Greater Manchester for helping take this work to the real world, Immense for calibrating the simulation model, and InnovateUK for funding that made it possible.
This work has been submitted to the IEEE for possible publication. Copyright may be transferred without notice, after which this version may no longer be accessible.
\IEEEtriggeratref{11}


%

\end{document}